\documentclass[letterpaper, 10 pt, conference]{ieeeconf}  

\IEEEoverridecommandlockouts                              
\usepackage{amsmath}
\usepackage{amssymb}
\usepackage{amsfonts}
\usepackage[pdftex]{graphicx}
\DeclareGraphicsExtensions{.pdf, .png, .jpg}
\usepackage{algorithm}
\usepackage[noend]{algpseudocode}
\algrenewcommand\algorithmicindent{0.7em}%
\usepackage{gensymb}
\usepackage[caption=false]{subfig}
\usepackage{cite}

\newcommand{\current}[1]{\mathbf{w}(#1)}

\newcommand{\Currentsimple}[0]{\mathbf{W}}
\newcommand{\stream}[1]{\phi(#1)}
\newcommand{\pos}[1]{\mathbf{x}_{#1}}
\newcommand{\Pos}[1]{\mathbf{X}_{#1}}
\newcommand{\poshat}[1]{ \mathbf{\hat{x}}_{#1} } 
\newcommand{\Poshat}[1]{ \mathbf{\hat{X}}_{#1} }
\newcommand{\posobs}[1]{ \mathbf{y}_{#1} }
\newcommand{\poserr}[1]{ \mathbf{\epsilon}_{#1} }
\newcommand{\vel}[1]{\mathbf{v}_{#1}}

\newcommand{\dt}{\Delta t}
\newcommand{\Likelihood}{\mathcal{P}(\Currentsimple_{k} \mid \mathbf{C}_{k}, \mathbf{W}_{1:k-1} )}
\DeclareMathOperator*{\argmax}{arg\,max}

\usepackage{amsthm} 
\newtheorem{problem}{Problem}
\newtheorem{assumption}{Assumption}

\theoremstyle{definition}

\title{\LARGE \bf
Online Estimation of Ocean Current from Sparse GPS Data for Underwater Vehicles
}

\author{Ki Myung Brian Lee$^1$, Chanyeol Yoo$^1$, Ben Hollings$^2$, Stuart Anstee$^3$, Shoudong Huang$^1$ and Robert Fitch$^1$ 
\thanks{This work is supported by an Australian Government Research Training Program (RTP) Scholarship, Australia's Defence Science and Technology Group, Blue Ocean Monitoring, and the University of Technology Sydney.}
\thanks{$^1$Authors are with the University of Technology Sydney, Ultimo, NSW 2006, Australia {\tt\footnotesize brian.lee@student.uts.edu.au, \{chanyeol.yoo, shoudong.huang, rfitch\}@uts.edu.au}}
\thanks{$^2$Author is with Blue Ocean Monitoring Ltd, Subiaco, WA 6008, Australia {\tt\footnotesize ben.hollings@blueoceanmonitoring.com}}
\thanks{$^3$Author is with the Defence Science and Technology Group, Department of Defence, Australia {\tt\footnotesize stuart.anstee@dst.defence.gov.au}}
}

\begin{document}

\maketitle
\thispagestyle{empty}
\pagestyle{empty}

\begin{abstract}
Underwater robots are subject to position drift due to the effect of ocean currents and the lack of accurate localisation while submerged. We are interested in exploiting such position drift to estimate the ocean current in the surrounding area, thereby assisting navigation and planning. We present a Gaussian process~(GP)-based expectation-maximisation~(EM) algorithm that estimates the underlying ocean current using sparse GPS data obtained on the surface and dead-reckoned position estimates. We first develop a specialised GP regression scheme that exploits the incompressibility of ocean currents to counteract the underdetermined nature of the problem. We then use the proposed regression scheme in an EM algorithm that estimates the best-fitting ocean current in between each GPS fix. The proposed algorithm is validated in simulation and on a real dataset, and is shown to be capable of reconstructing the underlying ocean current field. We expect to use this algorithm to close the loop between planning and estimation for underwater navigation in unknown ocean currents.  
\end{abstract}

\section{INTRODUCTION}

Ocean monitoring offers tremendous economic value with various applications such as oceanographic research~\cite{Rudnick2004}, military surveillance~\cite{Johannsson2010}, and oil and gas source localisation \cite{Russell-Cargill2018}. Various autonomous platforms have been used in such ocean monitoring tasks, including autonomous underwater vehicles (AUVs)~\cite{Stokey2005}, underwater gliders~\cite{Webb2001}, and even passive platforms without any actuation~\cite{Argo2000}. 

\begin{figure}[t]
    \centering
    \includegraphics[width=0.9\columnwidth]{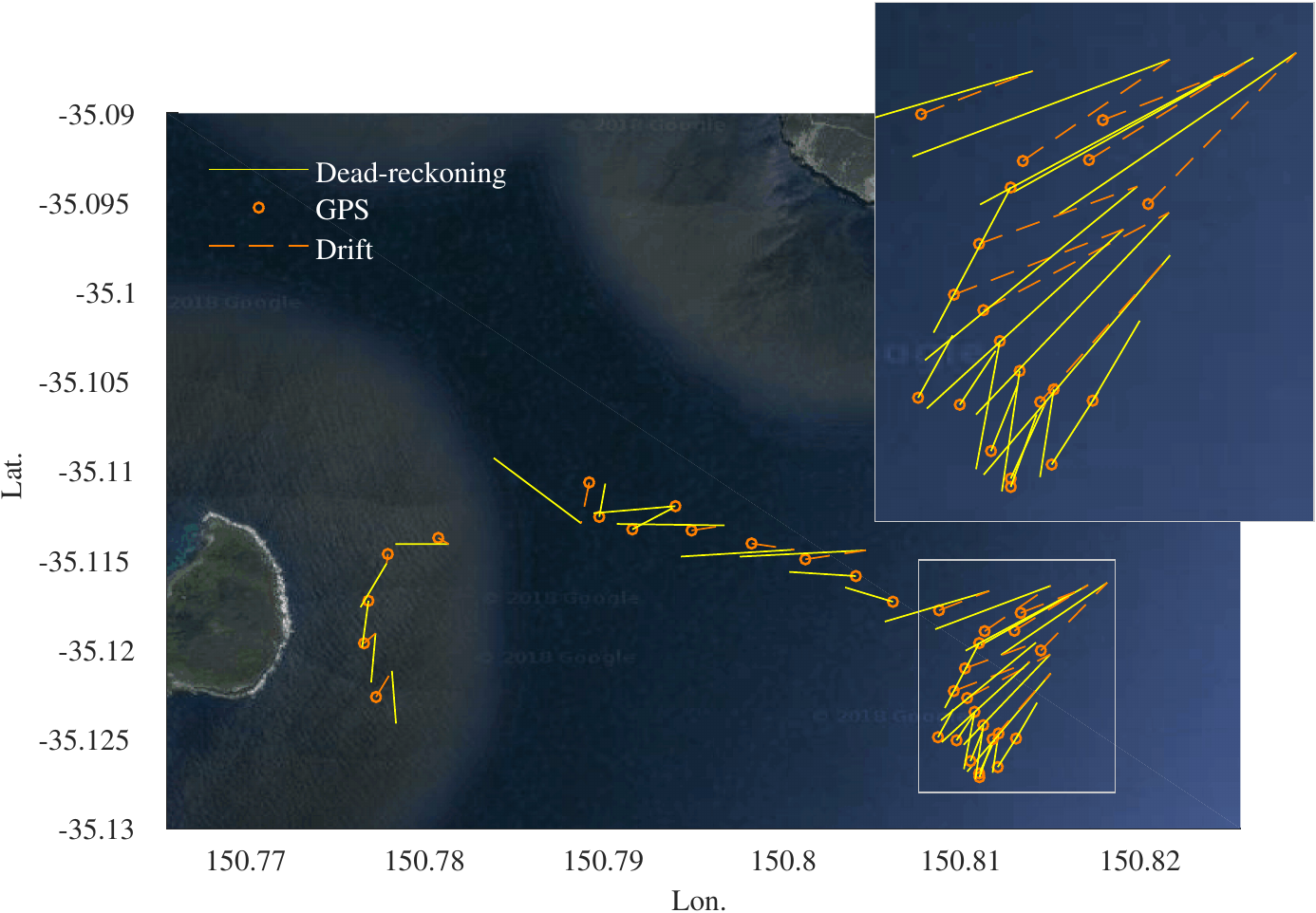}
    \vspace{-2ex}
    \caption{Result of a Slocum G3 underwater glider operation in Jervis Bay, Australia. The dead-reckoned position (yellow solid line) differ substantially from GPS measurements on surface (orange circle) due to ambient current. Orange dashed line shows the drift, which we use for estimating ocean current. Inset: trajectory inside the white box.\label{fig:jervis_bay_drift}} 
\end{figure}

A main challenge in underwater robotics is the effect of ambient ocean current. Due to lack of GPS while submerged, ocean currents can cause considerable position drift. This has a strong impact on the utility of the gathered data, as most ocean monitoring tasks concern spatial phenomena~\cite{Paull2018a}. Thus, there has been substantial work on navigation and planning in flow fields~\cite{Lee2015,Kularatne2018b,Hollinger2015,yoo2016online}. However, most of this work assumes that the flow field is given \emph{a priori}, e.g. from an external database~\cite{Oke2005, Oke2013, Shchepetkin2005}. Unfortunately, the spatiotemporal resolution or accuracy provided by most databases is insufficient for the purpose of navigation, as noted in~\cite{Hollinger2015}. 

We are interested in estimating the ocean current online, without any prior information. This idea was recently explored as a deterministic, discrete estimation problem based on a prior oceanic model~\cite{Chang2017} augmented by the \emph{drift} of the vehicle.
Here, we also exploit drift as a source of information, but focus on a continuous, probabilistic form. Drift is measured between the true position, measured by GPS at the surface, and the dead-reckoned trajectory. As GPS is unavailable underwater, the dead-reckoning estimate drifts substantially from the true trajectory. In the case of underwater gliders, this disparity is typically in the order of a few hundred metres, as shown in Figs.~\ref{fig:jervis_bay_drift} and~\ref{fig:perth_drift}.  

We propose an expectation-maximisation~(EM) algorithm for estimating ocean current given GPS measurements and a dead-reckoned trajectory. The problem is severely underdetermined as the position drift is the \emph{sum} of current along the trajectory. We present a Gaussian process (GP) regression technique that incorporates the concept of \emph{incompressibilty} to provide a physically meaningful constraint. 
Our algorithm is demonstrated both in simulation and using two experimental datasets collected by underwater gliders. The simulation results show that the algorithm is capable of estimating accurately, starting from a uniform prior on current. With the experimental datasets, we could not compare to a baseline due to lack of other sources of data at the site, but the estimated current aligns with observed drift and also aligns with the shoreline, which is an expected pattern. The significance of this result is that we can now attempt to close the loop between estimation and planning, hence enabling underwater navigation in unknown ocean current. 

\section{RELATED WORK}\label{sec:rel_works} 

To estimate ocean currents online, one approach is to consider current as a low-frequency disturbance and then apply an extended Kalman filter~(EKF)~\cite{Medagoda2016c} or nonlinear observer~\cite{Fan2016} in conjunction with acoustic sensors. However, modelling current as a temporal phenomenon clearly overlooks its spatial structure, and acoustic sensors typically require a stationary reference (e.g., the seabed)~\cite{Paull2014a}. 

An approach that does consider the spatial nature of the problem is presented in~\cite{Merckelbach2008}. The authors examine the feasibility of ocean current estimation through simply calculating the average current velocity by dividing the position drift by time. Unsurprisingly, the estimate is increasingly unreliable as distance between diving and surfacing locations grows, and no predictive capability is provided. 

An improvement to this concept was presented in~\cite{Chang2017}. The work proposes the `motion tomography' algorithm, which reconstructs the local ocean current from GPS measurements using techniques from the computer tomography~(CT) literature. The algorithm represents the ocean as a discrete grid and iteratively solves a linear system based on a prior obtained from an oceanic model, such as~\cite{Shchepetkin2005}. The authors in~\cite{Cho2016a} develop a similar technique with dense GPS measurements. In comparison, the algorithm we propose in this paper does not require prior information other than sparse GPS measurements. Because we use a GP, we can incorporate prior current measurements at any location, if available, and estimate current at any location with associated uncertainty. Further, our proposed method accounts for the incompressibility of the ocean current, which not only enhances the physical fidelity, but also aids with the quality of the estimates as will be shown later. Using a GP also aligns better with planning, as evident in previous work such as~\cite{Lui2016, Hollinger2015, Lee2015}. The techniques we use in this paper are in line with the Bayesian system identification literature~\cite{Turner2009a}, and are inspired by its rigorous theoretical analysis. 

\section{PROBLEM FORMULATION}\label{sec:problem_formulation} 
Suppose we have a continuous-time dynamic model of an underwater vehicle: 
\begin{equation}
    \dot{\mathbf{x}}_{t} = \mathbf{v}_{t} + \mathbf{w}(\mathbf{x}_{t})
    ,
\end{equation}
where~$\pos{t} \in \mathbb{R}^2$ is the position of the vehicle at time~$t$, $\vel{t} \in \mathbb{R}^2$ is the velocity through water at time~$t$, $\mathbf{w} \in \mathcal{C}^{\infty}(\mathbb{R}^{2})$ is the ocean current modelled as a smooth 2D vector field. For simplicity, we do not take into account the vertical motion of the vehicle. The continuous-time model is discretised as: 
\begin{equation}\label{eq:dynmodel} 
	\pos{t+1} = \pos{t} + (\vel{t} + \current{\pos{t}}) \dt
	,
\end{equation}
where $\dt$ is the sampling time. 

\begin{figure}[tp]
    \centering
    \includegraphics[width=0.9\columnwidth]{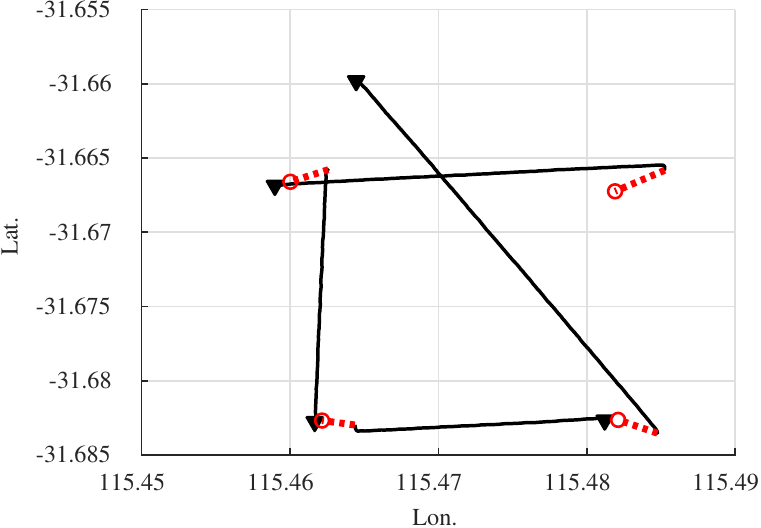}
    \caption{Data from Slocum G3 glider operation near Perth, Australia. Black bottom-facing triangles: dive-in points. Black solid line: dead-reckoned estimates ($\hat{\mathbf{x}}_{t}$). Red circles: GPS measurements ($\mathbf{y}_{\tau_{k}}$). Red dashed lines: the drift ($\Delta \mathbf{x}_{\tau_{k}}$) }  \label{fig:perth_drift}
\end{figure} 

The vehicle's velocity through water $\vel{t}$ is known, whereas the current~$\mathbf{w}(\pos{})$ is unknown. The vehicle estimates its own position based on \emph{dead-reckoning} assuming zero ocean current while submerged. The initial estimate derives from the last GPS measurement prior to dive-in. Namely,
\begin{equation}\label{eq:deadreckon} 
    \hat{\mathbf{x}}_{t+1} = \hat{\mathbf{x}}_t + \mathbf{v}_t \dt
    .
\end{equation}
The vehicle attempts to reach a target point using its dead-reckoned position estimate. When the vehicle's estimate is within a pre-set tolerance range from the target point, the vehicle climbs up to the surface and updates its position using GPS. We denote the time of surfacing events as $\tau_{k} \in [1, T]$, $\tau_{k} < \tau_{k+1}$. The GPS measurements are assumed to have i.i.d. Gaussian measurement error. Namely, 
\begin{align}
	\posobs{\tau_{k}} &= \pos{\tau_{k}} + \poserr{k} \label{eq:measmodel} \\
	\poserr{k} &\sim \mathcal{N}(\mathbf{0}, \sigma^{2}_{y}I )
	.
\end{align}

For the periods in between each GPS measurement, we use shorthand notation $\Pos{k} = \Pos{\tau_{k-1}:\tau_{k}}$ for true trajectory, $\Poshat{k} = \Poshat{\tau_{k-1}:\tau_{k}}$ for dead-reckoned trajectory, $\Currentsimple_{k} = \{\current{\pos{\tau_{k-1}}}, \ldots \current{\pos{\tau_{k}}} \}$ for current along trajectory, and $\posobs{k} = \posobs{\tau_{k}}$ for GPS measurements respectively. 

As the dead-reckoned estimates do not take ocean current into account, there is a substantial disparity between the dead-reckoned estimate and the GPS measurement. Throughout the rest of the paper, we refer to this disparity as \emph{drift}, and denote it by $\Delta \pos{k}$. In other words: 
\begin{equation} 
    \Delta \pos{k} = \posobs{\tau_{k}} - \poshat{\tau_{k}}. 
\end{equation} 
We will use the concept of a~\emph{cycle} to describe the three behaviours: 1) dive-in, 2) manoeuvre and 3) surfacing, as depicted in Fig.~\ref{fig:perth_drift}. Namely, a cycle~$\mathbf{c}_{k} = \{\Delta \pos{k}, \Poshat{k}\}$ is a tuple containing the dead-reckoned trajectory, $\Poshat{k}$, and the measured drift $\Delta \pos{k} $.

Although the current~$\current{\pos{}}$ throughout each cycle is unknown, we know that the drift measurements are related to the current. The ultimate aim of this paper is to solve the following \emph{maximum a posteriori} (MAP) estimation problem.
\begin{problem} [Ocean current estimation]\label{problem:full}
        Suppose we have a sequence of cycles~$\mathbf{C}_{1:k} = \mathbf{c}_1\mathbf{c}_{2}\cdots\mathbf{c}_{k}$. Find an optimal estimate for ocean current~$\mathbf{w}^{*}(\mathbf{x})$ over the space of 2D smooth vector fields $\mathcal{C}^{\infty}(\mathbb{R}^2)$ that maximises the posterior probability: 
    \begin{equation}\label{eq:full_likelihood} 
        \mathbf{w}^{*}(\mathbf{x}) = \argmax_{ \mathbf{w}(\mathbf{x}) \in \mathcal{C}^{\infty}(\mathbb{R}^2) } \mathcal{P}( \mathbf{w}(\mathbf{x}) \mid \mathbf{C}_{1:k}).
    \end{equation}
\end{problem}
Intuitively, solving the MAP problem implies that we find the ocean current $\mathbf{w}(\mathbf{x})$ that is the best trade-off between fitting 1) the drift measurements, $\Delta \mathbf{x}_{k}$ and 2) a constraint on the general behaviour of ocean current, which we will discuss in Sec.~\ref{sec:gp_incompressibility}. The constraint is necessary because there are infinitely many possibilities of ocean current vectors that sum up to the drift measurement $\Delta \mathbf{x}_{k}$. In other words, the problem is \emph{underdetermined}.

Finding a direct solution to Problem~\ref{problem:full} is difficult because there can be infinitely many relations between drift measurements~$\Delta \mathbf{x}_{k}$ and ocean current vectors~$\mathbf{w}(\mathbf{x})$ over the trajectory of vehicle~$\mathbf{x}_{t}$. In this paper, we make the following assumption about how current vectors at different positions~$\mathbf{w}(\mathbf{x})$ and the drift measurements are related.
\begin{assumption}[Conditional independence]\label{ass:condind} 
    For all $\pos{} \in \mathbb{R}^{2}$ such that $\pos{} \neq \pos{t}$, $\mathbf{w}(\mathbf{x})$ is conditionally independent of $\mathbf{C}_{1:k}$ given the current along trajectory,~$\Currentsimple_{1:k} = \{\Currentsimple_{1}, \cdots, \Currentsimple_{k} \}$. In other words, $\mathbf{w}(\mathbf{x})$ is indirectly related to $\mathbf{C}_{1:k}$ through $\Currentsimple_{1:k}$.
\end{assumption}

With the assumption, the overall problem can be re-written in a form that reveals two sub-problems: 
\begin{equation} \label{eq:full_likelihood_decomposition} 
    \mathcal{P}( \mathbf{w}(\mathbf{x}) \mid \mathbf{C}_{1:k} ) = \int \mathcal{P}( \mathbf{w}(\mathbf{x}) \mid \Currentsimple_{1:k}) \mathcal{P}( \Currentsimple_{1:k} \mid \mathbf{C}_{1:k} ) d\Currentsimple_{1:k} 
    ,
\end{equation}
where the sub-problems are to 1) estimate the current at a remote location given the current along the trajectory (i.e., $\mathcal{P}( \mathbf{w}(\mathbf{x}) \mid \Currentsimple_{1:k})$), and 2) estimate the current experienced along the vehicle's trajectory given drift measurements (i.e., $\mathcal{P}( \Currentsimple_{1:k} \mid \mathbf{C}_{1:k} )$).


\section{GP REGRESSION OF INCOMPRESSIBLE FLOW FIELDS}\label{sec:gp_incompressibility}

In this section, we solve for Subproblem~1, where we estimate the oceanic flow at a query position given flow at other locations (i.e., $\mathcal{P}( \mathbf{w}(\mathbf{x}) \mid \Currentsimple_{1:k})$). We model the oceanic flow with a GP, and impose the assumption of incompressibility. We first introduce incompressibility and the concept of a \emph{streamfunction}, then exploit the properties of the streamfunction to derive an incompressible GP. Incompressibility also serves as a useful constraint for solving Subproblem~2. 
We demonstrate an example with real ocean dataset to illustrate that the \emph{incompressible GP} outperforms the standard for modelling ocean currents. 

\subsection{Incompressibility and Streamfunction}

In this work, we model the ocean current as a planar, time-invariant and incompressible flow field. Planarity implies that the ocean current has no $z$-component, which describes the horizontal stratification of oceanic flow well (see, e.g., \cite{Zika2012}).

A flow field is \emph{incompressible}~\cite{Pritchard2011} when
\begin{equation}\label{eq:incompressibility} 
\nabla \cdot \current{\mathbf{x}} = 0
,
\end{equation} 
where $\nabla \cdot$ is the divergence operator. Intuitively, incompressibility implies that `the amount of water coming into a point is equal to the amount exiting the area'. As we do not expect to see surplus or deficit of water entering an area in the ocean, incompressibility is an appropriate description.  

If a planar flow is incompressible, it can be represented by a \emph{streamfunction}~$\phi : \mathbb{R}^2 \rightarrow \mathbb{R}$. Given a streamfunction $\stream{\mathbf{x}}$, one can compute the current $\current{\pos{}}$ as:
\begin{equation}\label{eq:stream2current} 
    \current{\pos{}} = \displaystyle \begin{bmatrix} \frac{\partial \stream{\pos{}}}{\delta y} 
    & -\frac{\partial \stream{\pos{}}}{\delta x}
    \end{bmatrix}^{T}
    .
\end{equation}
\subsection{Streamfunction-based GP Representation}

In this section, we show how to enforce the incompressibility condition in a GP model using a streamfunction. 
First, consider a streamfunction modelled as a GP:
\begin{equation}
    \phi(\mathbf{x}) \sim GP(0, k(\| \mathbf{x} - \mathbf{x'} \|))
    ,
\end{equation}
where $k(\| \mathbf{x} - \mathbf{x'} \|)$ is a kernel function. 

Because the derivative of a streamfunction~$\phi$ is flow field~$\mathbf{w}$ as shown in~\eqref{eq:stream2current} and the derivative of a GP is another GP~\cite{Martens2017}, our flow field can be represented by a GP. In infinite-dimensional Bayesian estimation, derivative operators apply to functions as do matrices to vectors \cite{Sarkka2012a}. Recall that if $ Cov(\mathbf{A}) = \Sigma_{\mathbf{A}} $ for a vector-valued random variable $\mathbf{A}$, $ Cov( M \mathbf{A} ) = M \Sigma_{\mathbf{A}} M^{T}$ given a matrix $M$.  

The derivative operators can be written as $\mathcal{D} = \begin{bmatrix} \frac{\partial} {\partial y} & -\frac{\partial }{ \partial x} \end{bmatrix}^{T}$ and $\mathcal{D}' = \begin{bmatrix} \frac{\partial} {\partial y'} & -\frac{\partial }{ \partial x'} \end{bmatrix}$ for the function case, and the flow field is represented by:  
\begin{align}
    \current{\pos{}} &= \mathcal{D} \stream{\pos{}} \sim GP(\mathbf{0}, \mathbf{K}(\mathbf{x}, \mathbf{x'})) , \label{eq:current_gp}
\end{align}
where the kernel function~$\mathbf{K}$ is given by: 
\begin{align}
    \mathbf{K}(\mathbf{x}, \mathbf{x'}) &= \mathcal{D} k(\| \mathbf{x} - \mathbf{x'} \|) \mathcal{D}' \notag \\
                                        &= \begin{bmatrix} \frac{\partial^{2} k} {\partial y^{2} } & - \frac{\partial^{2} k} {\partial x \partial y} \\
                                                           -\frac{\partial^{2} k } {\partial x \partial y} & \frac{\partial^{2} k} {\partial x^{2}}
                                            \end{bmatrix} \label{eq:current_kernel}  
    .
\end{align}
From the first line to the second line in~\eqref{eq:current_kernel}, we used the stationarity of the kernel. It is important to note that~\eqref{eq:current_kernel} can be computed analytically given a choice of kernel for the streamfunction~\cite{Martens2017, Solak2002}.


Using the GP representation of flow field~$\mathbf{w}$ with the kernel function in~\eqref{eq:current_kernel}, we can predict a set of current vectors~$\mathbf{W}(\mathbf{X}^{Q}) = \begin{bmatrix} \current{\pos{1}^{Q}} & \ldots & \current{\pos{N}^{Q}} \end{bmatrix}$ given previous measurement data~$\mathbf{W}(\mathbf{X}^{D}) = \begin{bmatrix} \current{\pos{1}^{D}} & \ldots & \current{\pos{M}^{D}} \end{bmatrix}$.
The predictions are given as a set of normal random variables:
\begin{align} 
    &\mathcal{P}(\mathbf{W}(\mathbf{X}^{Q}) \mid \mathbf{W}(\mathbf{X}^{D})) = \mathcal{N}(\mathbf{\mu}(\mathbf{X}^{Q}), \mathbf{\Sigma}(\mathbf{X}^{Q})) \label{eq:gp_normal}
    ,
\end{align}
with mean and covariance: 
\begin{align}
    &\mathbf{\mu}(\mathbf{X}^{Q}) = \mathbf{K}_{DQ}^{T}K_{DD}^{-1}\mathbf{W}_{D} \label{eq:gp_mean} \\
    &\mathbf{\Sigma}(\mathbf{X}^{Q}) = \mathbf{K}_{QQ} - \mathbf{K}_{DQ}^{T}K_{DD}^{-1}\mathbf{K}_{DQ} \label{eq:gp_cov} 
    ,
\end{align} 
where the matrices $\mathbf{K}_{DD}^{(i,j)} = \begin{bmatrix} \mathbf{K}(\pos{i}^{D}, \pos{j}^{D})\end{bmatrix}$, $\mathbf{K}_{DQ}^{(i,j)}  = \begin{bmatrix} \mathbf{K}(\pos{i}^{D}, \pos{j}^{Q})\end{bmatrix}$, and $\mathbf{K}_{QQ}^{(i,j)}  = \begin{bmatrix} \mathbf{K}(\pos{i}^{Q}, \pos{j}^{Q})\end{bmatrix}$ are constructed blockwise. 

\begin{figure*}[t]
    \centering
    \subfloat[Ocean current data from the Bureau of Meteorology, Australia]{\includegraphics[width=0.24\textwidth]{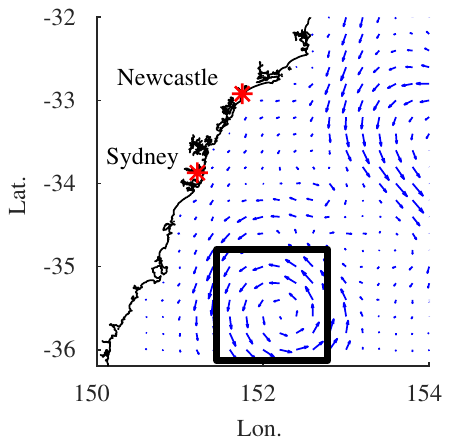}} \label{fig:gp_regression:current_data}
    \subfloat[True data inside square region, and the selected training data.]{\includegraphics[width=0.24\textwidth]{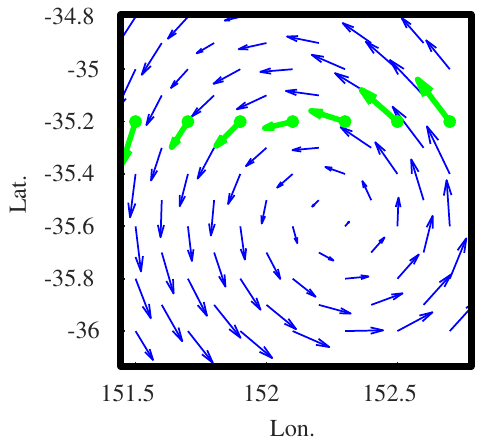} \label{fig:gp_regression:current_data_selected}} 
    \subfloat[Result with incompressible GP]{\includegraphics[width=0.24\textwidth]{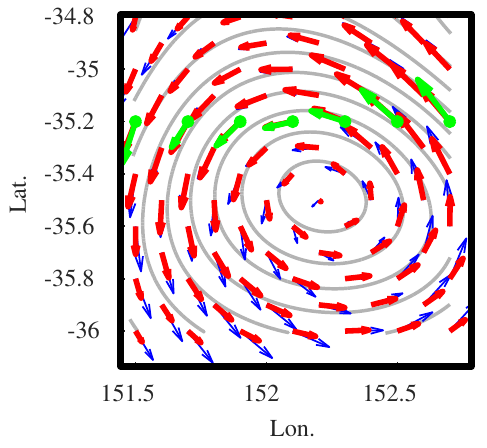} \label{fig:gp_regression:incompressible}}
    \subfloat[Result with standard GP]{\includegraphics[width=0.24\textwidth]{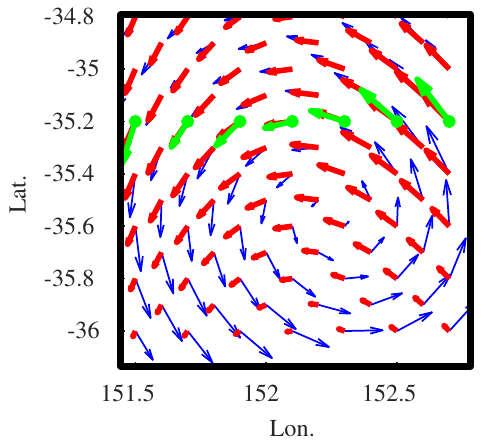} \label{fig:gp_regression:standard}}
    \caption{Comparison of the proposed and the standard multi-output GP. Blue: true data. Green: the training data used for regression. Red: estimated current. Gray: reconstructed streamline (only available with the proposed incompressible GP). Data were selected to emulate current estimated along a trajectory at each cycle. The proposed incompressible GP is capable of identifying large-scale eddy-like patterns, and hence offers better extrapolation. } 
    \label{fig:gp_regression}
\end{figure*}

\subsection{Comparison with Standard GP} 
We discuss how the incompressible GP is a better representation of the oceanic currents with a real dataset in Fig.~\ref{fig:gp_regression}. 
We selected a representative eddy from the east Australian current data provided by the Australian Bureau of Meteorology.
Then, we selected training samples along a line to emulate the current estimated along a trajectory. These training samples were extrapolated with GPs having the proposed incompressible kernel and the standard kernel $\mathbf{K}_{SE} = \mathbf{diag}(k_{SE}(\pos{}, \pos{}'), k_{SE}(\pos{}, \pos{}'))$. 

The standard kernel only fits a smooth vector field to the training samples. Meanwhile, the proposed incompressible GP extrapolates the ocean current much more accurately even with the limited training samples. An apparent benefit is that we can reconstruct eddy-like patterns~\cite{Oke2013}, which leads to a better extrapolation for the flow along a future trajectory given the estimate along the present trajectory.  

\section{EXPECTATION-MAXIMISATION FOR CURRENT ESTIMATION}\label{sec:leastsquares}

In this section, we solve Subproblem~2, which concerns estimating the current along trajectory. For simplicity, we will focus on estimating the flow along trajectory sequentially, given each incoming GPS measurements. In doing so, we are making a Markov assumption, where we fix the estimate of current along previous trajectories, $\Currentsimple_{1:k-1}$, when estimating the current along trajectory, $\Currentsimple_{k}$. It substantially reduces the computational complexity of the problem, as the algorithm is \emph{incremental}. More precisely, we assume: 
\begin{align} 
&\mathcal{P}( \Currentsimple_{1:k} \mid \mathbf{C}_{1:k} )\notag\\
= &\mathcal{P}( \Currentsimple_{k} \mid \mathbf{C}_{k}, \Currentsimple_{1:k-1} ) \mathcal{P}( \Currentsimple_{1:k-1} \mid \mathbf{C}_{1:k-1} )
,
\end{align}
which shows the problem reduces to estimating the current along present trajectory, given GPS measurements and previous current estimates (i.e. $\mathcal{P}( \Currentsimple_{k} \mid \mathbf{C}_{k}, \Currentsimple_{1:k-1} )$). 

The main challenge in estimating the current along trajectory arises from the strong causality between the trajectory itself and the current along the trajectory. To predict the current along trajectory, we must know the trajectory beforehand, and to predict the trajectory, we must know the current along trajectory. The challenge is solved through an EM algorithm. In each iteration, $i$, the EM algorithm iterates over estimating the trajectory $\Pos{k}^{i}$, called the \emph{expectation step} (E-step) and \ref{alg:incgp:em_estep}), and estimating the current along trajectory $\Currentsimple_{k}^{i}$ given $\Pos{k}^{i}$, called the \emph{maximisation step} (M-step). In effect, we iteratively `guess' the true trajectory, estimate the flow using the guess, and refine the guess on trajectory using the estimated flow. 

Subproblem~2 is re-written in the EM formulation:
\begin{align} 
    &\Likelihood \notag \\
    \propto &\int \mathcal{P}(\Currentsimple_{k} \mid \Pos{k}, \mathbf{C}_{k}, \Currentsimple_{1:k-1} ) \mathcal{P}(\Pos{k} \mid \Currentsimple_{k}, \mathbf{C}_{k} ) d\Pos{k} \notag \\
    = &\mathbb{E}_{\Pos{k} \mid \Currentsimple_{k}, \mathbf{C}_{k} }[\mathcal{P}(\Currentsimple_{k} \mid \Pos{k}, \mathbf{C}_{k}, \Currentsimple_{1:k-1} )] 
    \label{eq:posterior}
    .
\end{align} 

In the E-step (Sec.~\ref{subsec:e_step}), we find~$\mathcal{P}(\Pos{k}^{i} \mid \Currentsimple_{k}^{i-1}, \mathbf{C}_{k} )$ and evaluate the expectation~\eqref{eq:posterior}. In the M-step (Sec.~\ref{subsec:m_step}, we find
\begin{equation}\label{eq:em_posterior_max}
\mathbf{W}_{k}^{i} = \argmax_{\mathbf{W}} \mathbb{E}_{\Pos{k}^{i} \mid \Currentsimple_{k}^{i-1}, \mathbf{C}_{k} }[\mathcal{P}(\Currentsimple \mid \Pos{k}^{i}, \mathbf{C}_{k}, \Currentsimple_{1:k-1})].  
\end{equation} 

\subsection{E-step}\label{subsec:e_step} 

In the E-step, we need to compute the expectation in \eqref{eq:posterior}. In \eqref{eq:dynmodel}, notice that the only source of uncertainty derives from the fact that $\current{\pos{}}$ is a GP. Therefore, given current, the trajectory is fully known. Formally, the conditional distribution of the trajectory becomes a Dirac delta distribution: 
\begin{equation} 
    \mathcal{P}(\Pos{k}^{i} \mid \Currentsimple_{k}^{i-1}, \mathbf{C}_{k} ) = \delta( \Pos{k}^{i} - ( \Poshat{k} + B\Currentsimple_{k}^{i-1} ) ),
\end{equation}
where $B^{(i,j)} = \begin{bmatrix} \Delta t  \mathbf{1}_{2 \times 2} \end{bmatrix}$ if $i \leq j$, and $B^{(i,j)} = \begin{bmatrix} \mathbf{0}_{2 \times 2} \end{bmatrix}$ otherwise. $\mathbf{1}_{2 \times 2}$ and $\mathbf{0}_{2 \times 2}$ denote identity and zero matrices. 

As such, the expectation integral collapses to a mere evaluation at: 
\begin{equation}\label{eq:e_step_update} 
\Pos{k}^{i} = \Poshat{k} + B\Currentsimple_{k}^{i-1}
. 
\end{equation} 

\begin{figure*}[t]
    \centering
    \subfloat[Estimated flow field after 1 waypoint]{\includegraphics[width=0.67\columnwidth]{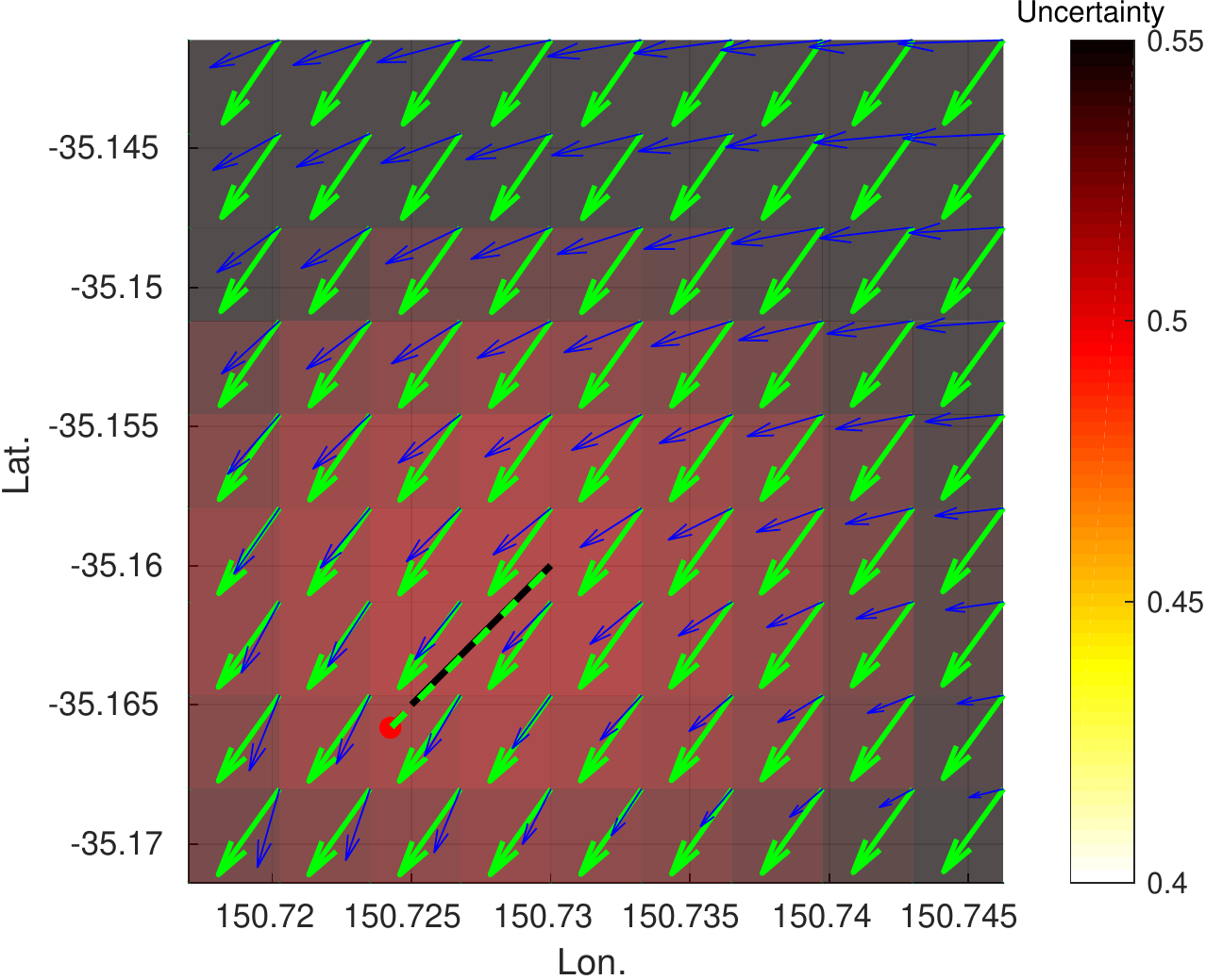} \label{fig:result_3}}
    \subfloat[Estimated flow field after 4 waypoints]{\includegraphics[width=0.67\columnwidth]{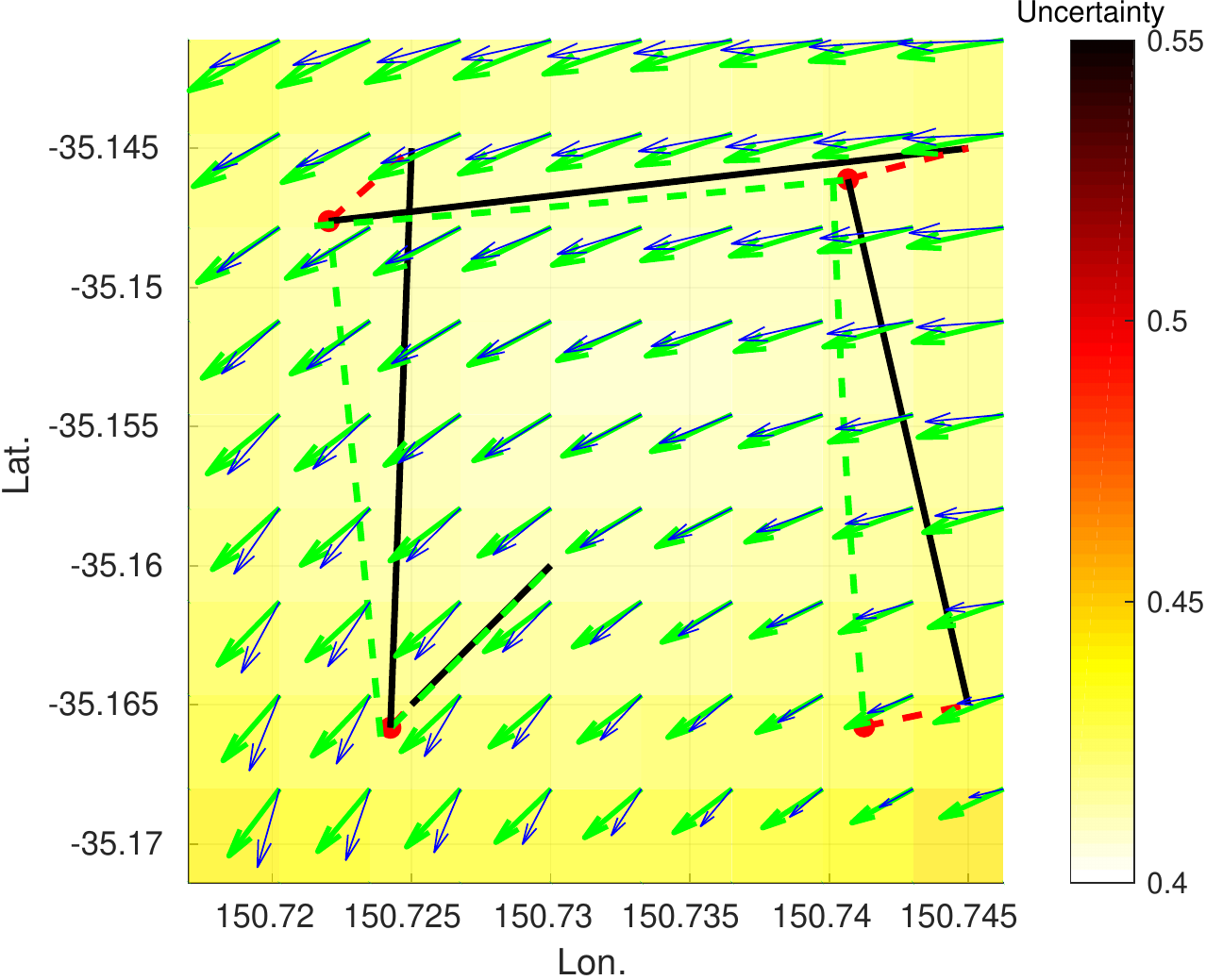} \label{fig:result_6}}
    \subfloat[Estimated flow field after 8 waypoints]{\includegraphics[width=0.67\columnwidth]{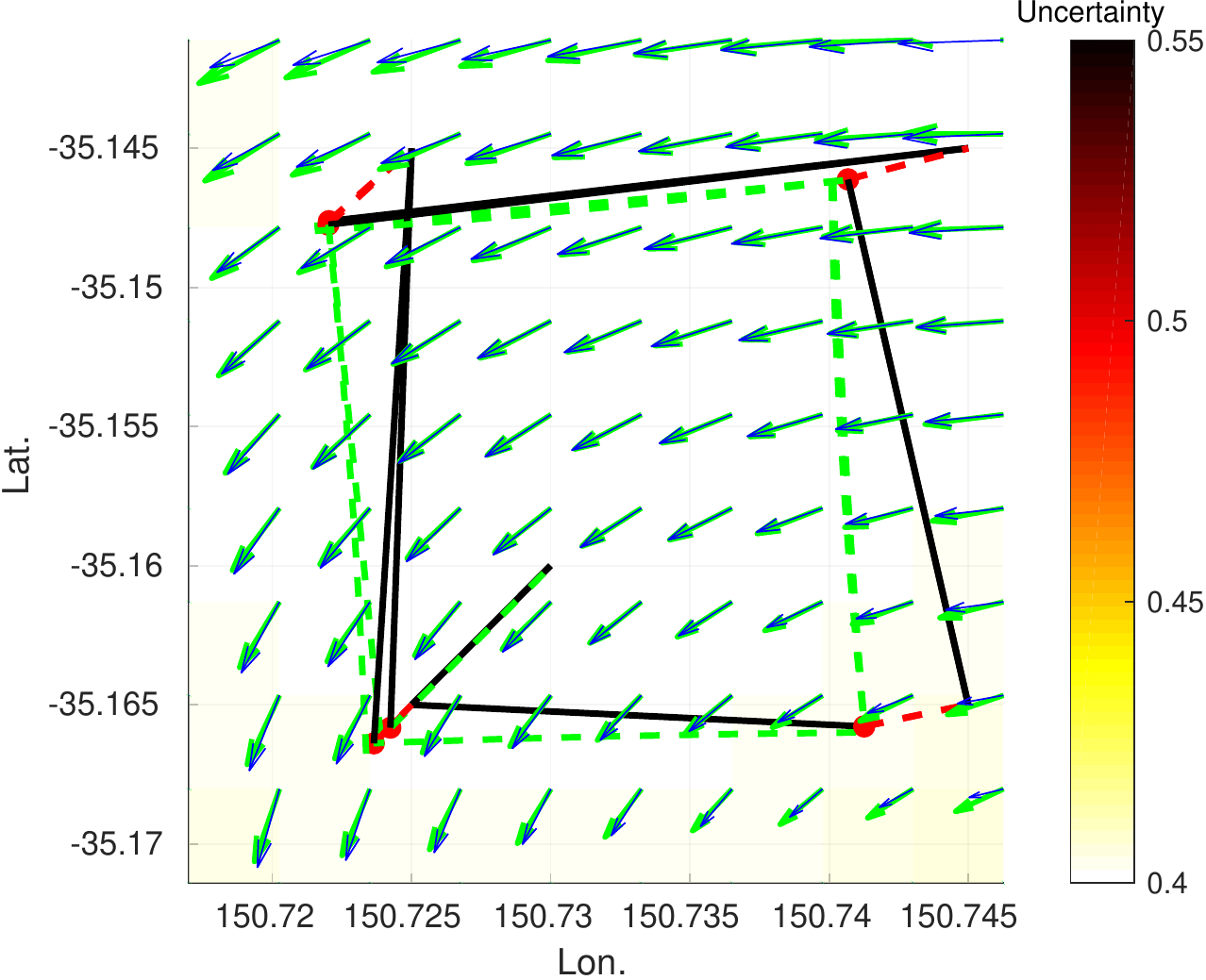} \label{fig:result_9}}
    \caption{Simulation results. Trajectory converted to lat-long for end-to-end testing. Black solid line: dead-reckoned trajectory. Green dashed line: reconstructed trajectory. Red markers: GPS. Red dashed line: drift. Blue arrows: true flow field. Green arrows: estimated flow field. Uncertainty refers to trace of covariance.}
    \label{fig:sim_results}
\end{figure*}
\begin{figure}[tp]
    \centering
    \includegraphics[height=0.22\textheight]{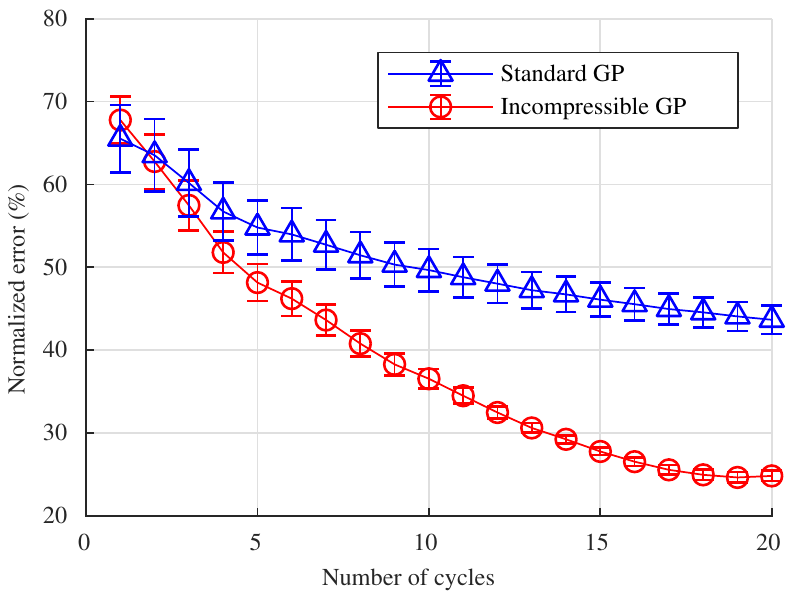}
    \caption{Convergence of Alg.~\ref{alg:incgp} with the standard multi-output GP (blue, triangle) and the incompressible GP (red, circle) for 100 different gyre patterns. Error normalized by the magnitude of the true current. 99\% confidence interval is shown.}
    \label{fig:convergence}
\end{figure}

\subsection{M-step}\label{subsec:m_step} 
In the M-step, we maximise the expectation taken in Sec.~\ref{subsec:e_step}. As shown in Sec.~\ref{subsec:e_step}, the expectation \eqref{eq:posterior} is a simple evaluation at~\eqref{eq:e_step_update}. As $\current{\pos{}}$ is a GP, this yields a prior given by a Gaussian random variable:   
\begin{equation}
    \mathbb{E}_{ \Pos{k}^{i} \mid \Currentsimple_{k}^{i-1}, \mathbf{C}_{k} } \left[ \mathcal{P}(\Currentsimple_{k}^{i} \mid \Pos{k}^{i}, \Currentsimple_{1:k-1} ) \right] = \mathcal{N}( \mathbf{\mu}(\Pos{k}^{i}), \mathbf{\Sigma}(\Pos{k}^{i}) ),
\end{equation} 
where $\mathbf{\mu}(\Pos{k}^{i})$ and $\mathbf{\Sigma}(\Pos{k}^{i})$ are calculated using GP prediction equations~\eqref{eq:gp_mean} and~\eqref{eq:gp_cov} given the current estimated with previous drift measurements, $\Currentsimple_{1:k-1}$. 

Notice that we can write $\Delta \pos{k}$ in terms of $\Currentsimple_{k}$ as: 
\begin{equation} 
    \Delta \pos{k} =  C \Currentsimple_{k} + \poserr{k} \label{eq:current2drift} 
    ,
\end{equation} 
where $C = \Delta t \begin{bmatrix} \mathbf{1}_{2 \times 2} & \mathbf{1}_{2 \times 2} & \ldots & \mathbf{1}_{2 \times 2} \end{bmatrix}$. Thus, $\Currentsimple_{k}^{i}$ and $\Delta \pos{k}$ are joint normal random variables: 
\begin{multline}\label{eq:joint_normal} 
    \mathbb{E}_{\Pos{k}^{i} \mid \Currentsimple_{k}^{i-1}, \mathbf{C}_{k} } \left[ \mathcal{P}(\Currentsimple_{k}^{i}, \Delta \pos{k} \mid \Pos{k}^{i}, \Currentsimple_{1:k-1}) \right] \\
  = \mathcal{N} \left( \begin{bmatrix} \mathbf{\mu}(\Pos{k}^{i}) \\ C \mathbf{\mu}(\Pos{k}^{i}) \end{bmatrix}, \begin{bmatrix} \Sigma(\Pos{k}^{i}) & \Sigma(\Pos{k}^{i}) C^{T} \\ C\Sigma(\Pos{k}^{i}) & C \Sigma(\Pos{k}^{i}) C^{T} + \sigma_{y}^{2} I_{2 \times 2} \end{bmatrix} \right)
    .
\end{multline} 

The merit of this formulation is that the maximising solution is now given in closed form, because maximising the posterior \eqref{eq:em_posterior_max} is equivalent to finding the conditional mean using \eqref{eq:joint_normal}. This is given by~\cite{Petersen2007}: 
\begin{equation}\label{eq:leastsquares-optimal}
    \begin{aligned} 
    \mathbf{W}_{k}^{i} = \mu + \Sigma C^{T} \left(C \Sigma C^{T} + \sigma_{y}^{2} I \right)^{-1} \left( \Delta \pos{T_{k}} - C\mu \right)
    \end{aligned} 
    ,
\end{equation}
where $\mathbf{\mu} = \mathbf{\mu}(\mathbf{X}_{k}^{i})$ and $\Sigma = \Sigma(\mathbf{X}_{k}^{i})$.  

\begin{algorithm} [t]
\caption{GP-EM algorithm for ocean current estimation}\label{alg:incgp}
\begin{algorithmic}[1]
\State $GP \gets \text{InitialiseEmptyGP}$\label{alg:incgp:init_gp} 
\While {Vehicle is operational}
\If {Drift measurement $\Delta \mathbf{x}_{k}$ available} 
\State Initialise $\Pos{k}^{0} \gets \Poshat{k}$\label{alg:incgp:em_start}
\For{i=1 \ldots N}
\State Update estimate of $\Currentsimple_{k}^{i}$ using~\eqref{eq:leastsquares-optimal} with~$\Pos{k}^{i-1}$\label{alg:incgp:em_estep}
\State Update estimate of $\Pos{k}^{i}$ using~\eqref{eq:e_step_update} with~$\Currentsimple_{k}^{i}$\label{alg:incgp:em_mstep}
\EndFor\label{alg:incgp:em_end} 
\State $GP \gets \text{UpdateGP}(GP, \Currentsimple_{k}^{N}, \Pos{k}^{N})$\label{alg:incgp:update_gp}
\EndIf
\EndWhile
\end{algorithmic}
\end{algorithm}

\subsection{Implementation} 


The algorithm for solving Subproblem~2 is shown in Alg.~\ref{alg:incgp}. 
We initialise the algorithm with a zero-mean GP without any measurements (Alg.~\ref{alg:incgp} line~\ref{alg:incgp:init_gp}). From lines~\ref{alg:incgp:em_start} to~\ref{alg:incgp:em_end}, we convert each incoming GPS measurement into a pair of true trajectory and the current along trajectory,~$\Pos{k}$ and $\Currentsimple_{k}$ using our EM algorithm. Afterwards, the current along trajectory is added to the measurement set of the GP, as `pseudo-target'~\cite{Turner2009a}. As more measurements become available, the GP produces better prior for the iteration. We found that the Markov assumption reduces the computation time substantially with minimal performance sacrifice. 

\section{RESULTS}\label{sec:results}

In this section, we present case studies of our algorithm with simulated example and real dataset collected from Teledyne Webb G3 Slocum gliders. We used the squared exponential kernel~\cite{Rasmussen2006} for the streamfunction, having lengthscale of $\ell = 35 km$, and self-variance of~$\sigma_{W}^{2} = 0.5 m^{2}s^{-2}$. The kernel for the current vectors~\eqref{eq:current_kernel} was computed analytically by taking the partial derivatives~\cite{Martens2017, Solak2002}. 
The hyperparameters were found by maximising the data likelihood of the dataset from the Australian Bureau of Meteorology using a standard hyperparameter learning procedure~\cite{Rasmussen2006}.

\subsection{Simulation Results}

We simulated an underwater vehicle travelling in a flow field, with true and dead-reckoned estimates evolving as~\eqref{eq:dynmodel} and~\eqref{eq:deadreckon}. The vehicle is given four waypoints, and surfaces when its dead-reckoned position estimate obtained using \eqref{eq:deadreckon} is within 100 metres from the current target waypoint. 
For the purpose of validation, we used a double-gyre model and ran the proposed algorithm. The results are shown in Fig.~\ref{fig:sim_results}. 



From Figs.~\ref{fig:result_3} to~\ref{fig:result_9}, it can be seen that the algorithm actively improves the estimate of current as the mission progresses. In Fig.~\ref{fig:result_3}, it can be seen that the estimate after only one cycle is as good as the average current method in~\cite{Merckelbach2008}. However, by the fourth cycle, it can be seen that the estimated and the true flow fields are already in good agreement, with minor disparity. By the eighth cycle, the estimated and the true flow fields are almost indistinguishable. The uncertainty of the estimated current also decreases. 

In order to examine the convergence of the proposed algorithm further, we performed a Monte Carlo simulation with 100 randomly generated double gyre patterns. We took the error between predicted and true currents on a selected grid, normalised by the magnitude of the true current. The convergence was also compared with a standard GP with kernel function $\mathbf{K}(\pos{}, \pos') = \mathbf{diag}(k_{SE}(\pos{}, \pos'), k_{SE}(\pos{}, \pos'))$, having identical parameters (i.e. no incompressibility assumption or streamfunction). The result is shown in Fig.~\ref{fig:convergence}.

Figure~\ref{fig:convergence} shows that the algorithm gradually learns any randomly generated flow field, which is demonstrated by the decrease in normalised error for both standard and the proposed incompressible GP. However, the incompressible GP shows a much faster rate of convergence and a lower steady-state error than the standard. 
The result clearly indicates that our incompressible GP outperforms over the standard in describing oceanic flows as shown in Fig.~\ref{fig:gp_regression}. 

\begin{figure}[tp]
    \centering
    \subfloat[Near Jervis Bay, Australia. Trajectory shown in Fig.~\ref{fig:jervis_bay_drift}.]{\includegraphics[width=0.95\columnwidth]{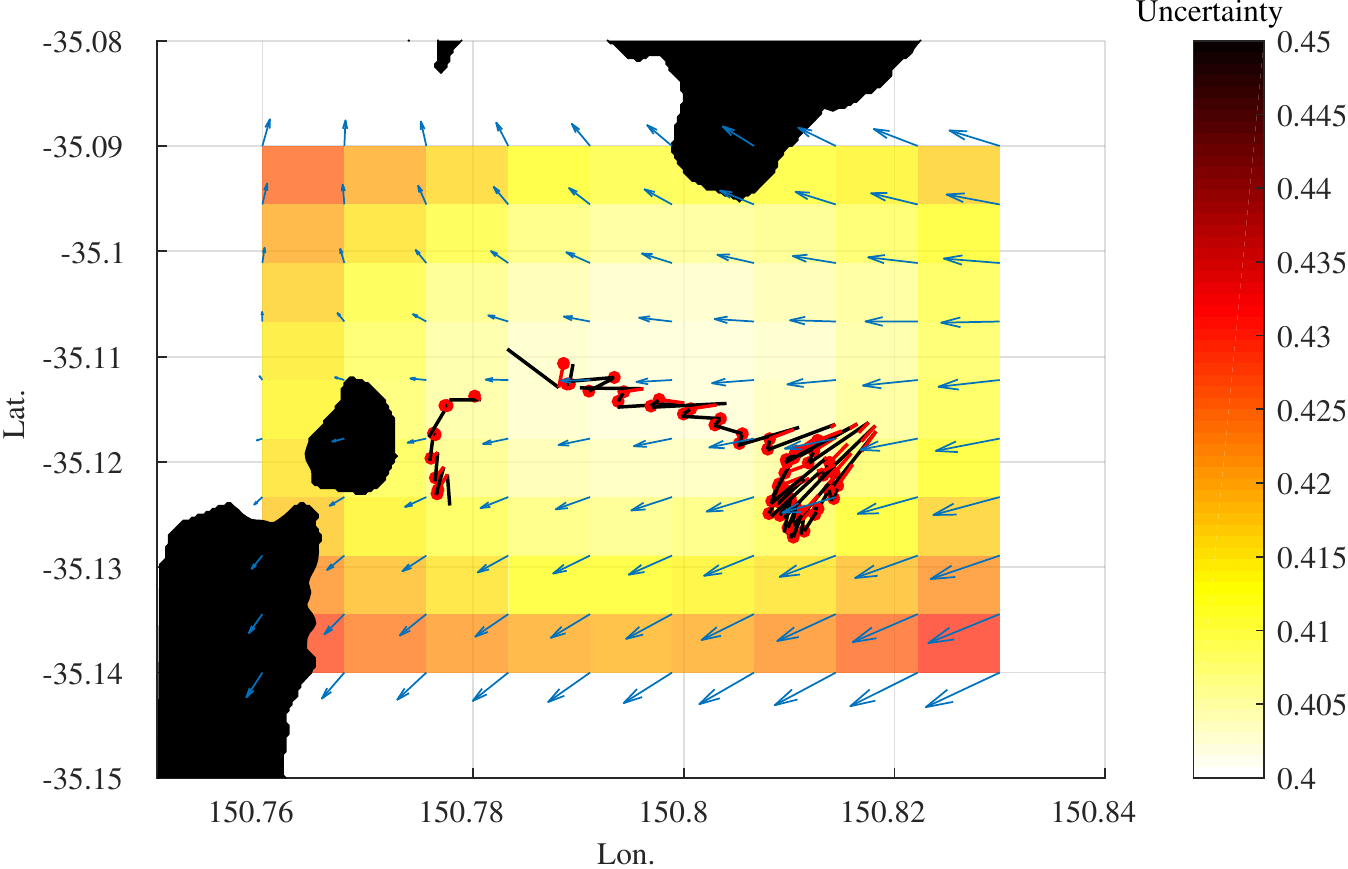} \label{fig:jervis_bay_result}} \\
    \subfloat[Near Perth, Australia. Trajectory shown in~Fig.~\ref{fig:perth_drift}.]{\includegraphics[width=0.95\columnwidth]{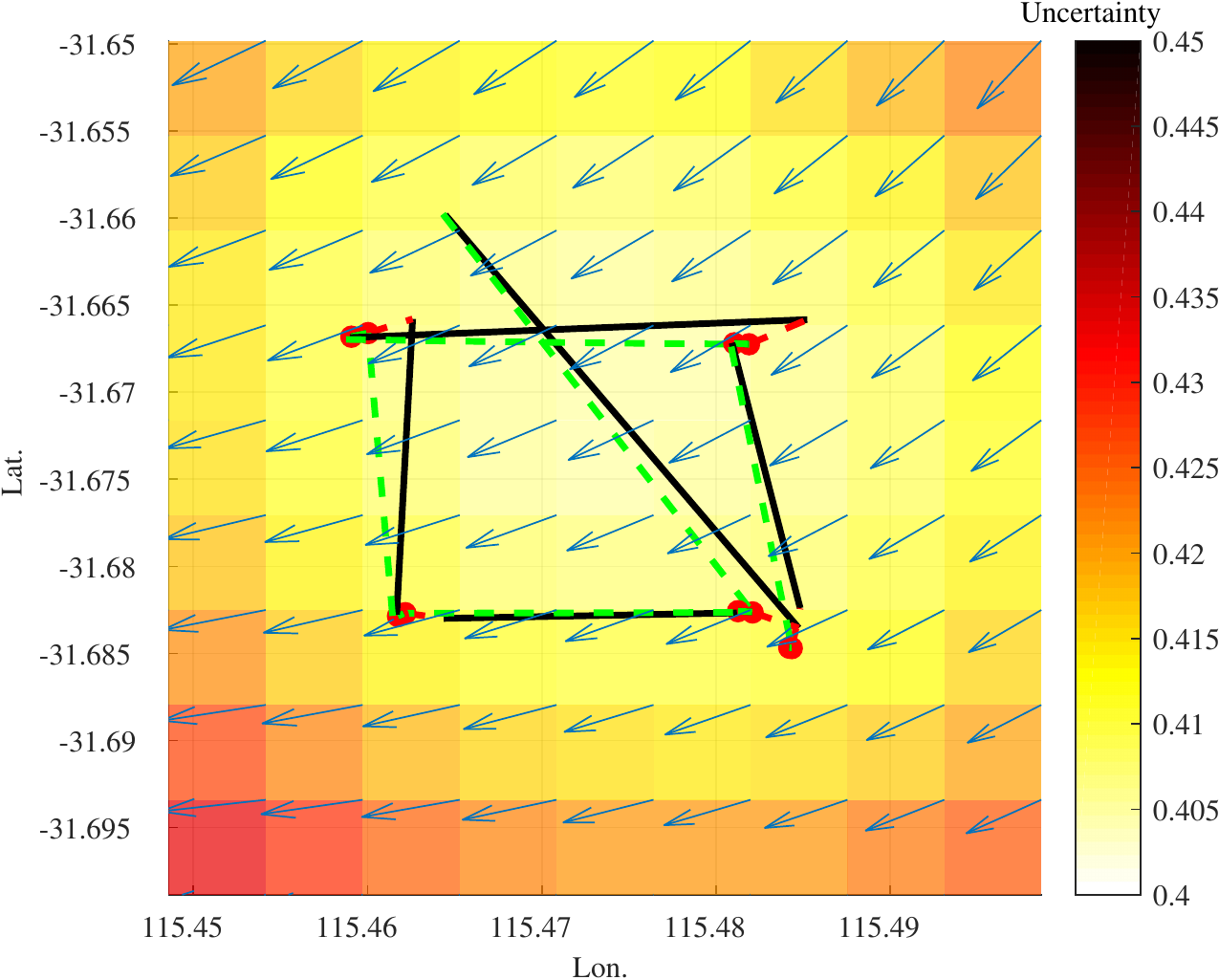} \label{fig:perth_result}}
    \caption{Field trial results showing estimated current (blue arrow), position from GPS (red circle), dead-reckoned path (black line) and estimated path (dashed green) }
    \label{fig:field_result} 
\end{figure}

\subsection{Field Results}

The proposed algorithm was tested in two field trials with a Slocum G3 underwater glider, one near Jervis Bay, Australia, and one in open ocean near Perth, Australia. The result from the Jervis Bay trial is shown in Fig.~\ref{fig:jervis_bay_result}, and the Perth trial in Fig.~\ref{fig:perth_result}. The glider was tasked to visit designated waypoints and communicate the current cycle~$\mathbf{c}_{k}$ when on surface. For the purpose of experiment, we disabled the onboard \emph{average current correction}~\cite{Merckelbach2008} to ensure the dead-reckoned estimate evolve as \eqref{eq:deadreckon}. 

For the Jervis Bay trial in Fig.~\ref{fig:jervis_bay_result}, it can be seen that the current estimated in Fig.~\ref{fig:jervis_bay_result} is in good agreement with what is expected near a bay: the ocean flow comes in from the open ocean, and majority of the flow enters the bay in alignment with the bay's shoreline. 
An important observation is that there exists only a small perpendicular ocean current to the land which well satisfies our common intuition.
The observation clearly indicates that our method accurately models the correlation between ocean currents at different points. This is because the method incorporates incompressibility, a physical attribute of the real ocean, unlike the standard GP-based model.


For the Perth trial in Fig.~\ref{fig:perth_result}, the predicted ocean current for Fig.~\ref{fig:perth_drift} within the operating zone had only one data point, pointing 45$\degree$ southeast. However, the prediction does not seem to agree with the drift observed. Meanwhile, the proposed algorithm can estimate a flow field that best explains the observed drift.

\section{Conclusion and Future Work}\label{sec:conc} 

We have developed an EM algorithm for estimating ocean current from sparse GPS data. The performance of the algorithm is improved by using a GP regression scheme that takes into account the physical intuition of incompressibility. The proposed algorithm was tested with both simulated and experimental data, with positive results. Our future work includes extending the algorithm to consider time- and depth-variation of ocean current, by adding extra dimensions to the kernel function. We would also like to combine the proposed algorithm with motion and task planning algorithms~\cite{yoo2013provably,yoo2016online}, and validate experimentally the performance of the whole framework. 

\newpage
\bibliography{GP_Ocean,generic}

\end{document}